\documentclass[journal]{IEEEtran}
\usepackage{hyperref}
\usepackage{url}
\usepackage{bm}
\usepackage{amsmath,amsfonts,amsthm} 
\usepackage{cite}
\usepackage{caption}
\usepackage{threeparttable}
\usepackage{graphicx}

\ifCLASSINFOpdf
\else
\fi
\begin{document}
%
\title{Spatio-Temporal Backpropagation for Training High-performance Spiking Neural Networks}
%
%
%

\author{Yujie Wu $^{\dag}$,
        Lei Deng $^{\dag}$,
        Guoqi Li,
        Jun Zhu  and
        Luping Shi 
\thanks{$^{\dag}$The authors contribute equally.}
\thanks{Yujie Wu, Lei Deng, Guoqi Li and Luping Shi are with Center for Brain-Inspired Computing Research (CBICR),Department of Precision Instrument,Tsinghua University, 100084 Beijing, China. (email:lpshi@tsinghua.edu.cn;)}
\thanks{Jun Zhu is with state Key Lab of Intelligence Technology and System,Tsinghua National Lab for Information Science and Technology,Tsinghua University, 100084 Beijing, China. (email:dcszj@mail.tsinghua.edu.cn)}
}

\maketitle
\begin{abstract}
 Compared with artificial neural networks (ANNs), spiking neural networks (SNNs) are promising to explore the brain-like behaviors since the spikes could encode more spatio-temporal information. Although existing schemes including pre-training from ANN or direct training based on backpropagation (BP) make the supervised training of SNNs possible, these methods only exploit the networks' spatial domain information which leads to the performance bottleneck and requires many complicated training techniques. Another fundamental issue is that the spike activity is naturally non-differentiable which causes great difficulties in training SNNs. To this end, we build an iterative LIF model that is friendlier for gradient descent training. By simultaneously considering the layer-by-layer spatial domain (SD) and the timing-dependent temporal domain (TD) in the training phase, as well as an approximated derivative for the spike activity, we propose a spatio-temporal backpropagation (STBP) training framework without using any complicated skill.  We design the corresponding fully connected and convolution architecture and evaluate our framework on the static MNIST and a custom object detection dataset, as well as the dynamic N-MNIST. Results show that our approach achieves the best accuracy compared with existing state-of-the-art algorithms  on spiking networks. This work provides a new perspective to explore the high-performance SNNs for future brain-like computing paradigm with rich spatio-temporal dynamics.
\end{abstract}
%

%
\IEEEpeerreviewmaketitle

\section{Introduction}

Deep neural networks (DNNs) have achieved outstanding performance in diverse areas \cite{Chaudhari2017Progressive,Deng2014Deep,Jia2014Caffe,Hinton2012Deep,He2014Spatial},
while it seems that the brain uses another network architecture, spiking neural networks, to realize various complicated cognitive functions\cite{Zhang2013Spike, Allen2009Cognitive, Kasabov2015Spiking}. Compared with the existing DNNs, SNNs mainly have two superiorities: 1) the spike pattern flowing through SNNs fundamentally codes more spatio-temporal information, while most DNNs lack timing  dynamics, especially the widely used feedforward DNNs; and 2) event-driven paradigm of SNNs can  make it more hardware friendly, and be adopted by many neuromorphic platforms \cite{Benjamin2014Neurogrid,Merolla2014Artificial,furber2014spinnaker,Hwu2016A,Esser2016Convolutional,Shi}.

However, it remains challenging in training SNNs because of the quite complicated dynamics and non-differentiable nature of the spike activity. In summary, there exist three kinds of training methods for SNNs: 1) unsupervised learning; 2) indirect supervised learning; 3) direct supervised learning. The first one origins from the biological synaptic plasticity for weight modification, such as spike timing dependent plasticity (STDP) \cite{Diehl2015Unsupervised, Querlioz2013Immunity,Kheradpisheh2016Bio}. Because it only considers the local neuronal activities, it is difficult to achieve high performance. The second one firstly trains an ANN, and then transforms it into its SNN version with the same network structure where the spiking rate of SNN neurons acts as the analog activity of ANN neurons \cite{Perezcarrasco2013Mapping, Diehl2015Fast,Peter2013Real,Hunsberger2015Spiking}.  This is not a bio-plausible way to explore the learning nature of SNNs.
The most promising method to obtain high-performance training is the recent direct supervised learning based on the gradient descent theory with error backpropagation. However,  such a method  only considers  the layer-by-layer spatial domain and ignores the dynamics in temporal domain \cite{O2016Deep,Lee2016Training}. Therefore many  complicated  training skills are required to improve  performance\cite{Diehl2015Fast,Neil2016Phased,Lee2016Training}, such as fixed-amount-proportional reset, lateral inhibition, error normalization, weight/threshold regularization, etc.  Thus, a more general dynamic model and learning framework on SNNs are highly required.

In this paper, we propose a direct supervised learning framework for SNNs which combines both the SD and TD in the training phase. Firstly, we build an iterative LIF model with SNNs dynamics but it is friendly for gradient descent training. Then we consider both the spatial direction and temporal direction during the error backpropagation procedure, i.e, spatio-temporal backpropagation (STBP), which significantly improves the network accuracy. Furthermore, we introduce an approximated derivative to address the non-differentiable issue of the spike activity. We test our SNNs framework by using the fully connected and convolution architecture on the static MNIST and a custom object detection dataset, as well as the dynamic N-MNIST. Many complicated training skills which are generally required by existing schemes, can be avoided due to the fact that our proposed method can make full use of STD information that captures the nature of SNNs. Experimental results show that our proposed method could achieve the best accuracy on either static or dynamic dataset, compared with existing state-of-the-art algorithms. The influence of TD dynamics and different methods for the derivative approximation are systematically analyzed. This work shall open a way to explore the high-performance SNNs for future brain-like computing paradigms with rich STD dynamics.
\section{Method and Material}
\begin{figure*}[!htb]
\includegraphics[width=10cm]{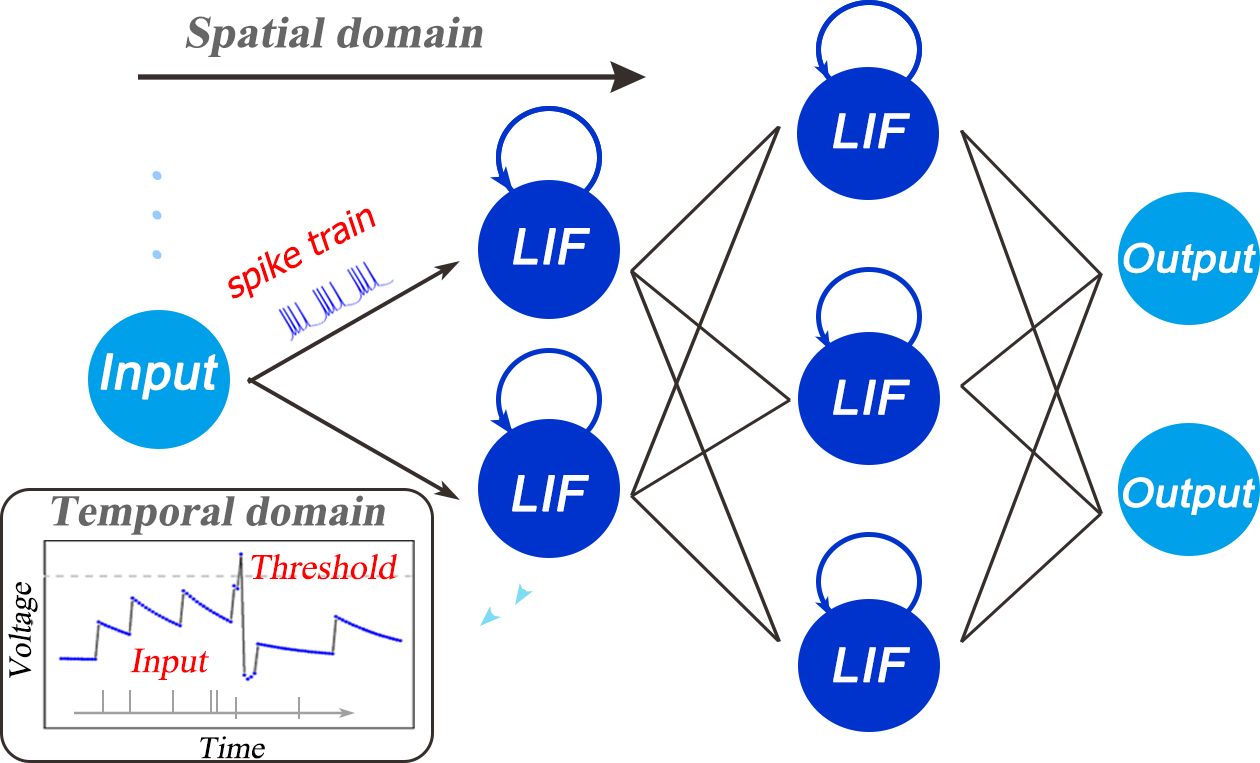}
\centering
\caption{\textbf{Illustration of the spatio-temporal characteristic of SNNs}. Besides the layer-by-layer spatial dataflow like ANNs, SNNs are famous for the rich temporal dynamics and non-volatile potential integration. However, the existing training algorithms only consider either the spatial domain such as the supervised ones via backpropagation, or the temporal domain such as the unsupervised ones via timing-based plasticity, which causes the performance bottleneck. Therefore, how to build an learning framework making full use of the spatio-temporal domain (STD) is fundamentally required for high-performance SNNs that forms the main motivation of this work.}
\label{fig1}
\end{figure*}
\subsection{Iterative Leaky Integrate-and-Fire Model in Spiking Neural Networks}
\label{section:2.2}
Compared with  existing deep neural networks, spiking neural networks  fundamentally code  more spatio-temporal information due to two  facts  that   i) SNNs   can   also have deep architectures like DNNs, and ii) each  neuron   has its own   neuronal dynamic properties.   The  former one   grants  SNNs   rich    spatial domain information while the later one  offers  SNNs the power of encoding   temporal domain  information.  However, currently there is no unified framework that allows   the effective training   of SNNs  just  as implementing   backpropagation (BP) in DNNs by considering the spatio-temporal dynamics. This  has challenged  the extensive use  of SNNs in various applications. In this work, we will present  a framework based on  iterative  leaky integrate-and-fire (LIF)   model   that enables us to  apply spatio-temporal  backpropagation for  training  spiking  neural  networks.

It is known that  LIF  is  the most widely  applied  model  to describe the  neuronal dynamics in SNNs,  and it can be simply governed by
\begin{align}
\tau \frac{du(t)}{dt} = -u(t) + I(t)
\label{LIF_1}
\end{align}
where $u(t)$ is the neuronal membrane potential at time $t$, $\tau$ is a time constant and $I(t)$  denotes  the pre-synaptic input which is determined by the pre-neuronal activities or external injections and the synaptic weights. When the membrane potential $u$ exceeds a given threshold $V_{th}$, the neuron fires a spike and resets its potential to $u_{reset}$.  As shown in Figure \ref{fig1},  the forward dataflow of the SNN propagates in the layer-by-layer SD like   DNNs, and the self-feedback injection at each neuron node generates non-volatile integration in the TD. In this way, the whole SNN runs with complex STD dynamics and codes spatio-temporal information into the spike pattern. The existing training algorithms only consider either the SD such as the supervised ones via backpropagation, or the TD such as the unsupervised ones via timing-based plasticity, which causes the performance bottleneck. Therefore, how to build an learning framework making full use of the STD is fundamentally required for high-performance SNNs that forms the main motivation of this work.

However,   obtaining  the analytic solution of LIF model in (\ref{LIF_1}) directly   makes it inconvenient/obscure to train SNNs based on backpropagation.
This is because the whole network shall  present complex dynamics in both SD and TD. To  address this issue,   the following       event-driven iterative  updating  rule
\begin{align}
u(t) = u(t_{i-1})e^{\frac{t_{i-1}-t}{\tau}}+I(t)
\label{LIF_2}
\end{align}
can be well  used to  approximate   the neuronal potential $u(t)$ in (\ref{LIF_1}) based on  the last spiking moment $t_{i-1}$ and the    pre-synaptic input $I(t)$.   The membrane potential exponentially decays until the neuron receives pre-synaptic inputs, and a new update round will start once the neuron fires a spike. That is to say,    the neuronal states are co-determined by the spatial accumulations of $I(t)$ and the leaky temporal memory of $u(t_{i-1})$.

As we know,  the efficiency of error backpropagation  for training DNNs greatly benefits from the iterative representation of gradient descent which yields the chain rule for layer-by-layer error propagation in the SD backward pass.
 This motivates us to  propose a iterative LIF   based SNN  in which the iterations occur in both the SD and TD as follows:
\begin{align}
\label{LIF_31}
x_i^{t+1,n} &= \sum_{j=1}^{l(n-1)}w_{ij}^{n} o_j^{t+1,n-1}\\
\label{LIF_32}
u_i^{t+1,n} &= u_i^{t,n}f(o_i^{t,n})+x_i^{t+1,n}+ b_i^{n}\\
\label{LIF_33}
o_i^{t+1,n} &= g(u_i^{t+1,n})
\end{align}
where
\begin{align}
 f(x)  &= \tau e^{-\frac{x}{\tau} }
 \label{fx}
\end{align}
\begin{eqnarray}
g(x) =
\begin{cases}
   1,  &x \geq V_{th} \\
    0,   &x <V_{th}
\end{cases}
 \label{gx}
\end{eqnarray}
In above formulas,  the upper index $t$ denotes the moment at time $t$,  and $n$ and $l(n)$  denote  the $nth$ layer and the number of neurons in the $nth$ layer, respectively.  $w_{ij}$ is the synaptic weight from the $jth$ neuron in pre-synaptic layer to the $ith$ neuron in the post-synaptic layer, and  $o_j\in \{0, 1\}$ is the neuronal output of the $jth$ neuron where $o_j=1$ denotes a spike activity and $o_j=0$ denotes nothing occurs.  $x_i$ is a simplified representation of the pre-synaptic inputs of the $ith$ neuron, similar to the $I$ in the original LIF model. $u_i$ is the neuronal membrane potential of the $ith$ neuron and $b_i$ is a bias parameter related the threshold $V_{th}$.

Actually,  formulas (\ref{LIF_32})-(\ref{LIF_33}) are  also inspired from  the LSTM model \cite{Gers1999Learning,hochreiter1997long,Chung2015Gated}  by using a forget gate $f(.)$ to control the TD memory and an output gate $g(.)$    to fire a spike.   The forget gate $f(.)$  controls  the leaky extent of the potential memory in the TD,   the output gate  $g(.)$   generates a spike activity when it is activated.  Specifically,  for a   small positive time constant $\tau$, $f(.)$ can be approximated as
\begin{eqnarray}
f(o_i^{t,n})\approx
\begin{cases}
   \tau,  &o_i^{t,n}=0\\
  0,   &o_i^{t,n}=1
\end{cases}
\end{eqnarray}
since  $\tau e^{-\frac{1}{\tau}} \approx 0$.
In this way, the original LIF model could be transformed to an iterative version  where the recursive relationship in both the SD and TD is clearly describe, which is friendly for the following gradient descent training in the STD.

\subsection{Spatio-Temporal Backpropagation Training}
\label{section:2.3}
In order to present STBP training methodology,  we define the following loss function $L$   in which the  mean square error  for  all samples under  a given time windows $T$ is to be minimized
\begin{align}
L = \frac{1}{2S}\sum_{s=1}^S \parallel \bm{y_s} - \frac{1}{T} \sum_{t=1}^T \bm{o_s}^{t,N} \parallel_2^2
\label{loss}
\end{align}
where $\bm{y_s}$ and  $\bm{o_s}$  denote the label vector of the $s$th  training sample and the neuronal output vector  of the last layer $N$, respectively.
\begin{figure*} [!htb]
\centering
\includegraphics[width=12cm]{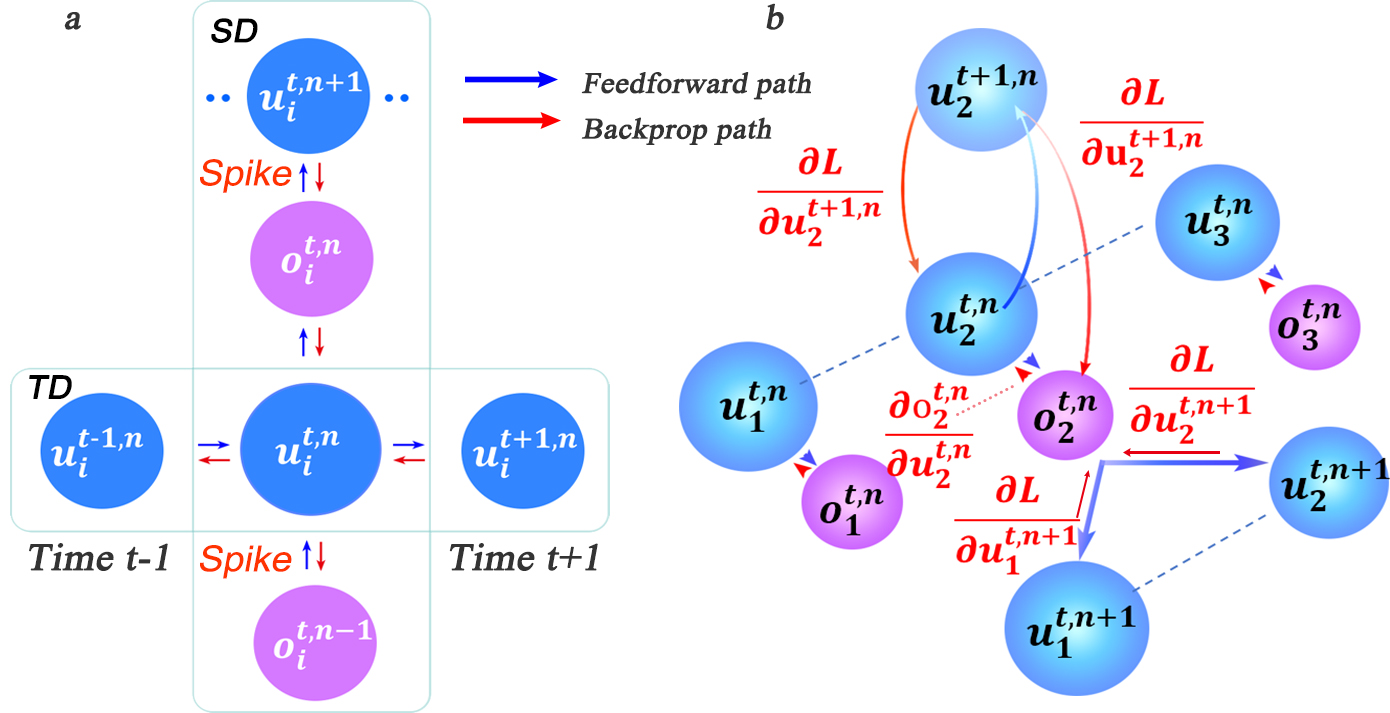}
\caption{\textbf{Error propagation in the STD.} (a)  At the single-neuron level, the vertical path and horizontal path represent the error propagation in the SD and TD, respectively. (b) Similar propagation occurs at the network level, where the error in the SD requires the multiply-accumulate operation like the feedforward computation.}
\label{fig2}
\end{figure*}

By combining equations (\ref{LIF_31})-(\ref{loss}) together it can be seen that $L$  is a function of $\bm{W}$ and $\bm{b}$. Thus, to obtain the derivative of $L$ with respect to $\bm{W}$ and $\bm{b}$ is required for the STBP algorithm based on gradient descent. Assume that we have obtained derivative of $\frac{\partial L}{\partial o_i }$ and $\frac{\partial L}{\partial u_i }$ at each layer $n$ at time $t$, which is an essential step to obtain the final $\frac{\partial L}{\partial \bm{W}}$ and   $\frac{\partial L}{\partial \bm{b}}$. Figure\ref{fig2} describes the error propagation (dependent on the derivation) in both the SD and TD at the single-neuron level (figure\ref{fig2}.a) and the network level (figure\ref{fig2}.b). At the single-neuron level, the propagation is decomposed into a vertical path of SD and a horizontal path of TD. The dataflow of error propagation in the SD is similar to the typical BP for DNNs, i.e. each neuron accumulates the weighted error signals from the upper layer and iteratively updates the parameters in different layers; while the dataflow in the TD shares the same neuronal states, which makes it quite complicated to directly obtain the analytical solution. To solve this problem, we use the proposed iterative LIF model to unfold the state space in both the SD and TD direction, thus the states in the TD at different time steps can be distinguished that enables the chain rule for iterative propagation. Similar idea can be found in the BPTT algorithm for training RNNs in \cite{Werbos1990Backpropagation}.

Now, we discuss how to obtain the complete gradient descent based on the following four cases. Firstly, we denote that
\begin{align}
\delta_i^{t,n} &= \frac{\partial L}{\partial o_i^{t,n} }
\end{align}
\textbf{Case 1:  $t=T$ at the output layer $n=N$.}\\
In this case, the derivative  $\frac{\partial L}{\partial o_i^{T,N}}$  can be directly obtained since it depends on the loss function in Eq.(\ref{loss}) of the output layer. We could have

\begin{align}
\label{case1-o}
\frac{\partial L}{\partial o_i^{T,N}}
&= -\frac{1}{TS}(y_i - \frac{1}{T}\sum_{k=1}^To_i^{k,N}).
\end{align}

The derivation with respect to $u_i^{T,N}$ is generated based on $o_i^{T,N}$

\begin{align}
\label{case1-u}
\frac{\partial L}{\partial u_i^{T,N}}
&= \frac{\partial L}{\partial o_i^{T,N}}\frac{{\partial o_i^{T,N}}}{\partial u_i^{T,N}}= \delta_i^{T,N}\frac{{\partial o_i^{T,N}}}{\partial u_i^{T,N}}.
\end{align}

\textbf{Case 2:  $t=T$ at the layers $n<N$.}\\
In this case, the derivative  $\frac{\partial L}{\partial o_i^{T,n}}$ iteratively depends on the error propagation in the SD at time $T$ as the typical BP algorithm. We have

\begin{align}
\frac{\partial L}{\partial o_i^{T,n}}
&=
\sum_{j=1}^{l(n+1)}\delta_j^{T,n+1}\frac{\partial o_j^{T,n+1}}{\partial o_i^{T,n}} =  \sum_{j=1}^{l(n+1)}\delta_j^{T,n+1} \frac{\partial{g}}{\partial u_i^{T,n}}w_{ji}.
\label{case2-o}
\end{align}

Similarly, the derivative  $\frac{\partial L}{\partial u_i^{T,n}}$ yields

\begin{align}
\label{case2p-u}
\frac{\partial L}{\partial u_i^{T,n}}
 &=
\frac{\partial L}{\partial u_i^{T+1,n}}\frac{\partial u_i^{T+1,n}}{\partial u_i^{T,n}} = \frac{\partial L}{\partial u_i^{T+1,n}}f(o_i^{T+1,n}).
 \end{align}

\textbf{Case 3:  $t< T$ at the output layer $n=N$.}\\
In this case, the derivative  $\frac{\partial L}{\partial o_i^{t,N}}$ depends on the error propagation in the TD direction. With the help of the proposed iterative LIF model in Eq.(\ref{LIF_31})-(\ref{LIF_33}) by unfolding the state space in the TD, we acquire the required derivative based on the chain rule in the TD as follows

\begin{align}
\label{case3-o}
\frac{\partial L}{\partial o_i^{t,N}}
&= \delta_i^{t+1,N}\frac{\partial o_i^{t+1,N}}{\partial o_i^{t,N}}+\frac{\partial L}{\partial o_i^{T,N}} \\
&= \delta_i^{t+1,N}\frac{\partial{g}}{\partial u_i^{t+1,N}}u_i^{t,N}
\frac{\partial f}{\partial o_j^{t,N}}+\frac{\partial L}{\partial o_i^{T,N}},
\end{align}

\begin{align}
\label{case3-u}
\frac{\partial L}{\partial u_i^{t,N}}
 &=
\frac{\partial L}{\partial o_i^{t,N}} \frac{\partial o_i^{t,N}}{\partial u_i^{t,N}}= \delta_i^{t,N}\frac{\partial g}{\partial u_i^{t,N}},
 \end{align}

where $\frac{\partial L}{\partial o_i^{T,N}} = -\frac{1 }{TS}(y_i - \frac{1}{T}\sum_{k=1}^To_i^{k,N})$ as in Eq.(\ref{case1-o}).

\textbf{Case 4:  $t<T$ at the layers $n<N$.}\\
In this case, the derivative $\frac{\partial L}{\partial o_i^{t,n}}$ depends on the error propagation in both SD and TD. On one side, each neuron accumulates the weighted error signals from the upper layer in the SD like Case 2; on the other side, each neuron also receives the propagated error from self-feedback dynamics in the TD by iteratively unfolding the state space based on the chain rule like Case 3. So we have

\begin{align}
\frac{\partial L}{\partial o_i^{t,n}}
&=
\sum_{j=1}^{l(n+1)}\delta_j^{t,n+1}\frac{\partial o_j^{t,n+1}}{\partial o_i^{t,n}} + \frac{\partial L}{\partial o_i^{t+1,n}}\frac{\partial o_i^{t+1,n}}{\partial o_i^{t,n}}\\
&=  \sum_{j=1}^{l(n+1)}\delta_j^{t,n+1} \frac{\partial{g}}{\partial u_i^{t,n}}w_{ji} +
\delta_i^{t+1,n}\frac{\partial{g}}{\partial u_i^{t,n}}u_i^{t,n}\frac{\partial f}{\partial o_i^{t,n}},\\
\frac{\partial L}{\partial u_i^{t,n}}
 &=
\frac{\partial L}{\partial o_i^{t,n}} \frac{\partial o_i^{t,n}}{\partial u_i^{t,n}}+\frac{\partial L}{\partial u_i^{t+1,n}}\frac{\partial u_i^{t+1,n}}{\partial u_i^{t,n}} \\
&= \delta_i^{t,n}\frac{\partial g}{\partial u_i^{t,n}} + \frac{\partial L}{\partial u_i^{t+1,n}}f(o_i^{t+1,n}).
\label{case4-o}
 \end{align}

Based on the four cases, the error propagation procedure (depending on the above derivatives) is shown in Figure\ref{fig2}. At the single-neuron level (Figure\ref{fig2}.a), the propagation is decomposed into the vertical path of SD and the horizontal path of TD. At the network level (Figure\ref{fig2}.b), the dataflow of error propagation in the SD is similar to the typical BP for DNNs, i.e. each neuron accumulates the weighted error signals from the upper layer and iteratively updates the parameters in different layers; and in the TD the neuronal states are unfolded iteratively in the timing direction that enables the chain-rule propagation.
Finally, we obtain the derivatives with respect to $\bm{W}$ and $\bm{b}$ as follows

\begin{align}
\label{udpate1}
\frac{\partial L}{\partial {\bm{b^n}}}
&= \sum_{t=1}^{T}\frac{\partial L}{\partial \bm{u^{t,n}}}\frac{\partial \bm{u^{t,n}}}{\bm{b^n}} =
\sum_{t=1}^{T}\frac{\partial L}{\partial \bm{u^{t,n}}},\\
\label{udpate2}
\frac{\partial L}{\partial \bm{W^n}}
&= \sum_{t=1}^{T}\frac{\partial L}{\partial \bm{u^{t,n}}}\frac{\partial \bm{u^{t,n}}}{\bm{W^n}} =
\sum_{t=1}^{T}\frac{\partial L}{\partial \bm{u^{t,n}}}{\bm{o^{t,n-1}}}^T,
\end{align}

where $\frac{\partial L}{\partial \bm{u^{t,n}}}$ can be obtained from in Eq.(\ref{case1-o})-(\ref{case4-o}). Given the $\bm{W}$ and $\bm{b}$ according to the STBP, we can use gradient descent optimization algorithms to effectively train SNNs for achieving high performance.

\subsection{Derivative Approximation of the Non-differentiable Spike Activity}
\begin{figure*}[!ht]
\centering
\includegraphics[width=9.5cm]{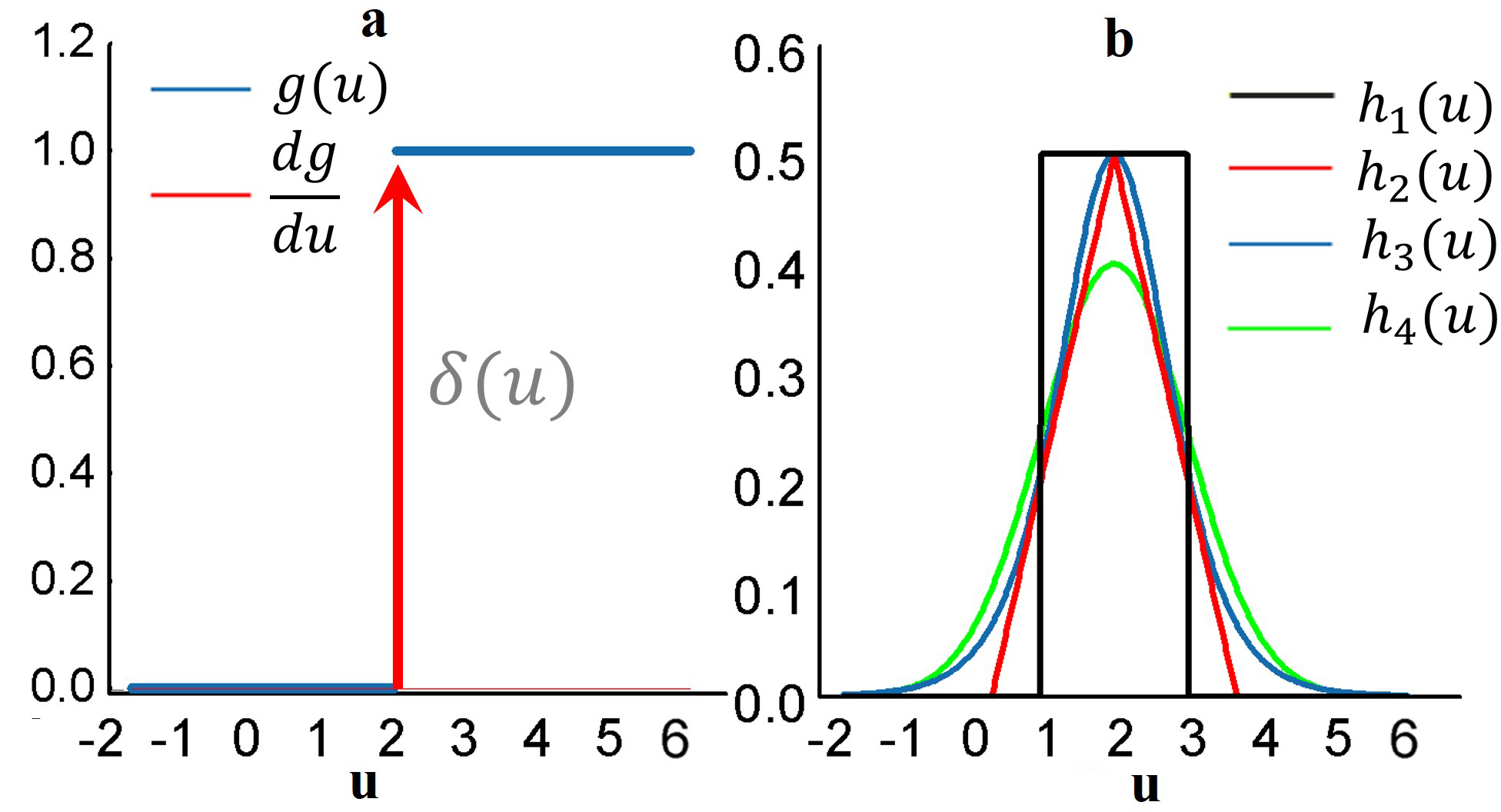}
\caption{\footnotesize{\textbf{Derivative approximation of the non-differentiable spike activity.} (a) Step activation function of the spike activity and its original derivative function which is a typical Diract function $\delta(u)$  with infinite value at $u=0$ and zero value at other points. This non-differentiable property disables the error propagation. (b)Several typical curves to approximate the derivative of spike activity. }}
\label{fig3}
\end{figure*}

In the previous sections, we  have presented  how to obtain the gradient information based on STBP,  but the issue of non-differentiable points at each spiking time is yet to be addressed.  Actually, the derivative of output gate $g(u)$ is required for the STBP training of Eq.(\ref{case1-o})-(\ref{udpate1}). Theoretically, $g(u)$ is a non-differentiable Dirac function of $\delta(u)$ which greatly challenges the effective learning of SNNs \cite{Lee2016Training}. $g(u)$ has zero value everywhere except an infinity value at zero, which causes the gradient vanishing or exploding issue that disables the error propagation. One of existing method viewed the discontinuous points of the potential at spiking times as noise and claimed it is beneficial for the model robustness\cite{Bengio2015An, Lee2016Training}, while it did not directly address the non-differentiability of the spike activity. To this end, we introduce four curves to approximate the derivative of spike activity denoted by $h_1$, $h_2$, $h_3$ and $h_4$ in Figure\ref{fig3}.b:

\begin{align}
\label{dg1}
h_1(u) &= \frac{1}{a_{1}}sign(|u-V_{th}|<\frac{a_{1}}{2}),\\
h_2(u) &= (\frac{\sqrt{a_{2}}}{2}-\frac{a_{2}}{4}|u-V_{th}|)sign(\frac{2}{\sqrt{a_{2}}}-|u-V_{th}|),\\
h_3(u) &= \frac{1}{a_{3}}\frac{e^{\frac{V_{th}-u}{a_{3}}}}{(1+e^{\frac{V_{th}-u}{a_{3}}})^2},\\
h_4(u) &= \frac{1}{\sqrt{2\pi a_{4}}}e^{-\frac{(u-V_{th})^2}{2a_{4}}},
\end{align}

where $a_i (i=1,2,3,4)$ determines the curve shape and steep degree. In fact, $h_1$, $h_2$, $h_3$ and  $h_4$ are the derivative of the rectangular function, polynomial function, sigmoid function and Gaussian cumulative distribution function, respectively. To be consistent with the Dirac function $\delta(u)$, we introduce the coefficient $a_i$ to ensure the integral of each function is 1. Obviously, it can be proven that  all the above  candidates satisfy that

\begin{align}
\lim \limits_{a_i\rightarrow0^{+}}h_i(u)=\frac{dg}{du},  i = 1,2,3,4.
\end{align}

 Thus, $\frac{ \partial g}{\partial u}$ in
Eq.(\ref{case1-o})-(\ref{udpate1}) for  STBP  can be approximated by
 \begin{align}
\frac{dg}{d u}\approx  \frac{d h_i}{d u}.
\end{align}
In section \ref{section-3.3}, we will analyze the influence on the SNNs performance with different curves and different values of $a_i$.

\section{Results}
\subsection{Parameter Initialization}
 The initialization of parameters, such as the weights, thresholds and other parameters, is crucial for stabilizing the firing activities of the whole network. We should simultaneously ensure timely response of pre-synaptic stimulus but avoid too much spikes that reduces the neuronal selectivity. As it is known that the multiply-accumulate operations of the pre-spikes and weights, and the threshold comparison are two key steps for the computation in the forward pass. This indicates the relative magnitude between the weights and thresholds determines the effectiveness of parameter initialization. In this paper, we fix the threshold to be constant in each neuron for simplification, and only adjust the weights to control the activity balance. Firstly, we initial all the weight parameters sampling from the standard uniform distribution
\begin{align}
\bm{W} \sim U[-1, 1]
\end{align}
Then, we normalize these parameters by
\begin{align}
w_{ij}^{n} &= \frac{w_{ij}^{n}}{\sqrt{\sum_{j=1}^{l(n-1)} {w_{ij}^{n}}^2}}, \quad   i= 1, .., l(n)
\end{align}
The set of other parameters is presented in Table\ref{table1}. Furthermore, throughout all the simulations in our work, any complex skill as in \cite{Diehl2015Fast,Lee2016Training} is no longer required, such as the fixed-amount-proportional reset, error normalization, weight/threshold regularization, etc. \\

\begin{table*}[!ht]
\centering
\caption{Parameters set in our experiments}
\label{table1}
{
\begin{tabular}{ccc}
\hline
Network parameter    & Description  & Value  \\
\hline
$T$                    & Time window                         & 30ms                    \\
$V_{th}$                    & Threshold (MNIST/object detection dataset/N-MNIST)            & 1.5, 2.0, 0.2                     \\
$\tau$               & Decay factor (MNIST/object detection dataset/N-MNIST) & 0.1ms, 0.15ms, 0.2ms                  \\
$a_{1},a_{2},a_{3},a_{4}$              & Derivative approximation parameters(Figure\ref{fig3})                      & 1.0                       \\
 \hline
$dt$                   & Simulation time step                                      & 1ms                     \\
$r$                    & Learning rate (SGD)                                 & 0.5                   \\
$\beta_{1}, \beta_{2}, \lambda$          & Adam parameters                                & 0.9, 0.999, 1-$10^{-8}$                    \\\hline
\end{tabular}}
\end{table*}

\subsection{Dataset Experiments}
\label{section:3.2}
We test our SNNs model and the STBP training method on various datasets, including the static MNIST and a custom object detection dataset, as well as the dynamic N-MNIST dataset. The input of the first layer should be a spike train, which requires us to convert the samples from the static datasets into spike events. To this end, the Bernoulli sampling from original pixel intensity to the spike rate is used in this paper.

\subsubsection{Spatio-temporal fully connected neural network}
\begin{figure*}[!ht]
\centering
\includegraphics[width=9.5cm]{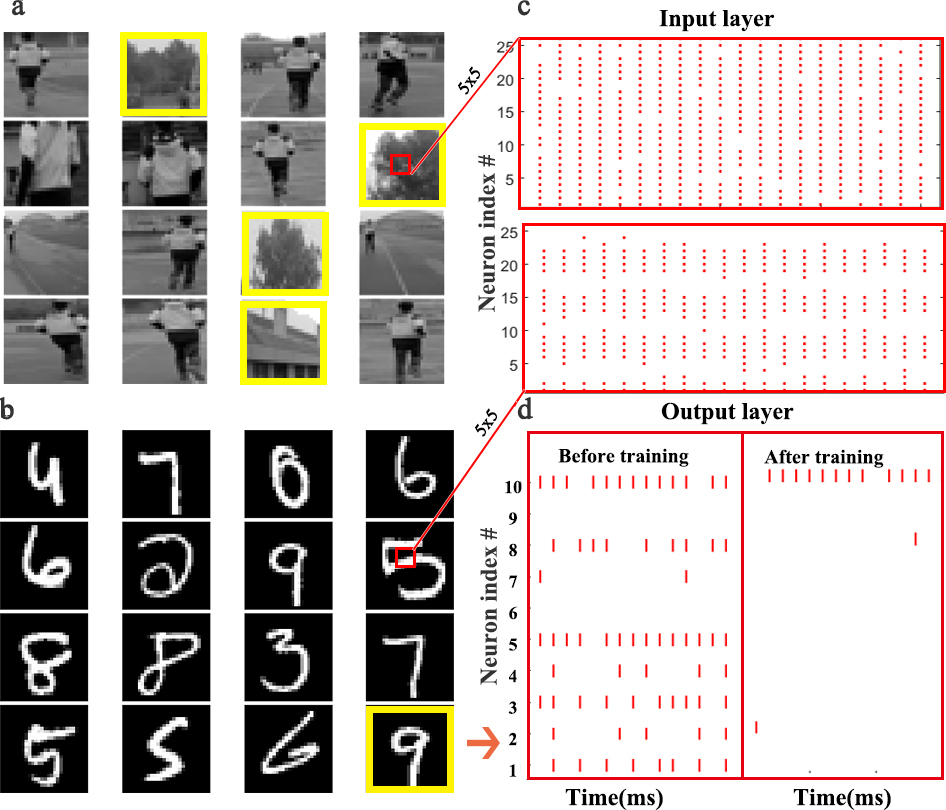}
\caption{\textbf{Static dataset experiments.} (a) A custom dataset for object detection. This dataset is a two-category image set built by our lab for pedestrian detection. By detecting whether there is a pedestrian, an image sample is labelled by 0 or 1. The images in the yellow boxes are labelled as 1, and the rest ones are marked as 0. (b)MNIST dataset. (c) Raster plot of the spike pattern of 49 input neurons converted from the center patch of 5$\times$5 pixels of a sample example on the object detection dataset (up) and MNIST (down). (d) Raster plot presents the comparison of output spike pattern before and after the STBP training over a digit 9 on MNIST dataset.}
\label{fig4}
\end{figure*}

\textbf{Static Dataset.} The MNIST dataset of handwritten digits \cite{L1998Gradient} (figure\ref{fig4}.b) and a custom dataset for object detection \cite{Shi} (figure\ref{fig4}.a) are chosen to test our method.
MNIST is comprised of a training set with 60,000 labelled hand-written digits, and a testing set of other 10,000 labelled digits, which are generated from the postal codes of 0-9. Each digit sample is a 28$\times$28 grayscale image. The object detection dataset is a two-category image dataset created by our lab for pedestrian detection. It includes 1509 training samples and 631 testing samples of 28$\times28$ grayscale image. By detecting whether there is a pedestrian, an image sample is labelled by 0 or 1, as illustrated in Figure\ref{fig4}.a. The upper and lower sub-figures in Figure\ref{fig4}.c are the spike pattern of 25 input neurons converted from the center patch of 5$\times$5 pixels of a sample example on the object detection dataset and MNIST, respectively. Figure\ref{fig4}.d illustrates an example for the spike pattern of output layer within 15ms before and after the STBP training over the stimulus of digit 9. At the beginning, neurons in the output layer randomly fires, while after the training the 10th neuron coding digit 9 fires most intensively that indicates correct inference is achieved.\\

\begin{table*}[!ht]
\centering
\caption{Comparison with the state-of-the-art spiking networks with similar architecture on MNIST.}
\begin{small}
\scalebox{0.98}{
\begin{tabular}{cccc}
\hline
Model                              & Network structure  & Training skills               & Accuracy       \\ \hline
Spiking RBM (STDP)\cite{Neftci2013Event} & 784-500-40                  & None                 & 93.16\%        \\
Spiking RBM(pre-training*)\cite{Peter2013Real}   & 784-500-500-10                  & None & 97.48\%        \\
Spiking MLP(pre-training*) \cite{Diehl2015Fast}          & 784-1200-1200-10 & Weight normalization & 98.64\%        \\
Spiking MLP(BP) \cite{O2016Deep}              & 784-200-200-10   & None                 & 97.66\%        \\
Spiking MLP(STDP) \cite{Diehl2015Unsupervised}  & 784-6400        & None                 & 95.00\%        \\
Spiking MLP(BP)  \cite{Lee2016Training}       & 784-800-10       & \begin{tabular}{c} Error normalization/\\parameter regularization  \end{tabular}& 98.71\%        \\
Spiking MLP(STBP)                       & 784-800-10       & None                 & \textbf{98.89\%} \\ \hline
\end{tabular}}
\end{small}
\begin{tablenotes}
\item[1] \footnotesize{We mainly compare with these methods that have the similar network architecture, and * means that their model is based on pre-trained ANN models.}
\end{tablenotes}
\label{table2}
\end{table*}

Table\ref{table2} compares our method with several other advanced results that use the similar MLP architecture on MNIST. Although we do not use any complex skill, the proposed STBP training method also outperforms all the reported results. We can achieve 98.89\% testing accuracy which performs the best. Table\ref{table3} compares our model with the typical MLP on the object detection dataset. The contrast model is one of the typical artificial neural networks (ANNs), i.e. not SNNs, and in the following we use 'non-spiking network'  to  distinguish them.
 It can be seen that our model achieves better performance than the non-spiking MLP. Note that the overall firing rate of the input spike train from the object detection dataset is higher than the one from MNIST dataset, so we increase its threshold to 2.0 in the simulation experiments.

\begin{table*}[!ht]
\begin{small}
\centering
\caption{Comparison with the typical MLP over object detection dataset.}
\label{table3}
\scalebox{1}{
\begin{tabular}{cccc}
\hline
Model     & Network structure & \multicolumn{2}{c}{Accuracy} \\ \cline{3-4}
          &                   & Mean   & Interval$^{*}$       \\ \hline
Non-spiking MLP(BP)       & 784-400-10               & 98.31\% & [97.62\%, 98.57\%]  \\
Spiking MLP(STBP) & 784-400-10     & \textbf{98.34\%} & [\textbf{97.94\%}, \textbf{98.57\%}] \\
\hline
\end{tabular}}
\begin{tablenotes}
\item[1] \footnotesize{* results with epochs [201,210].}
\end{tablenotes}
\end{small}
\end{table*}

\begin{figure}[!ht]
\centering
\includegraphics[width=9cm]{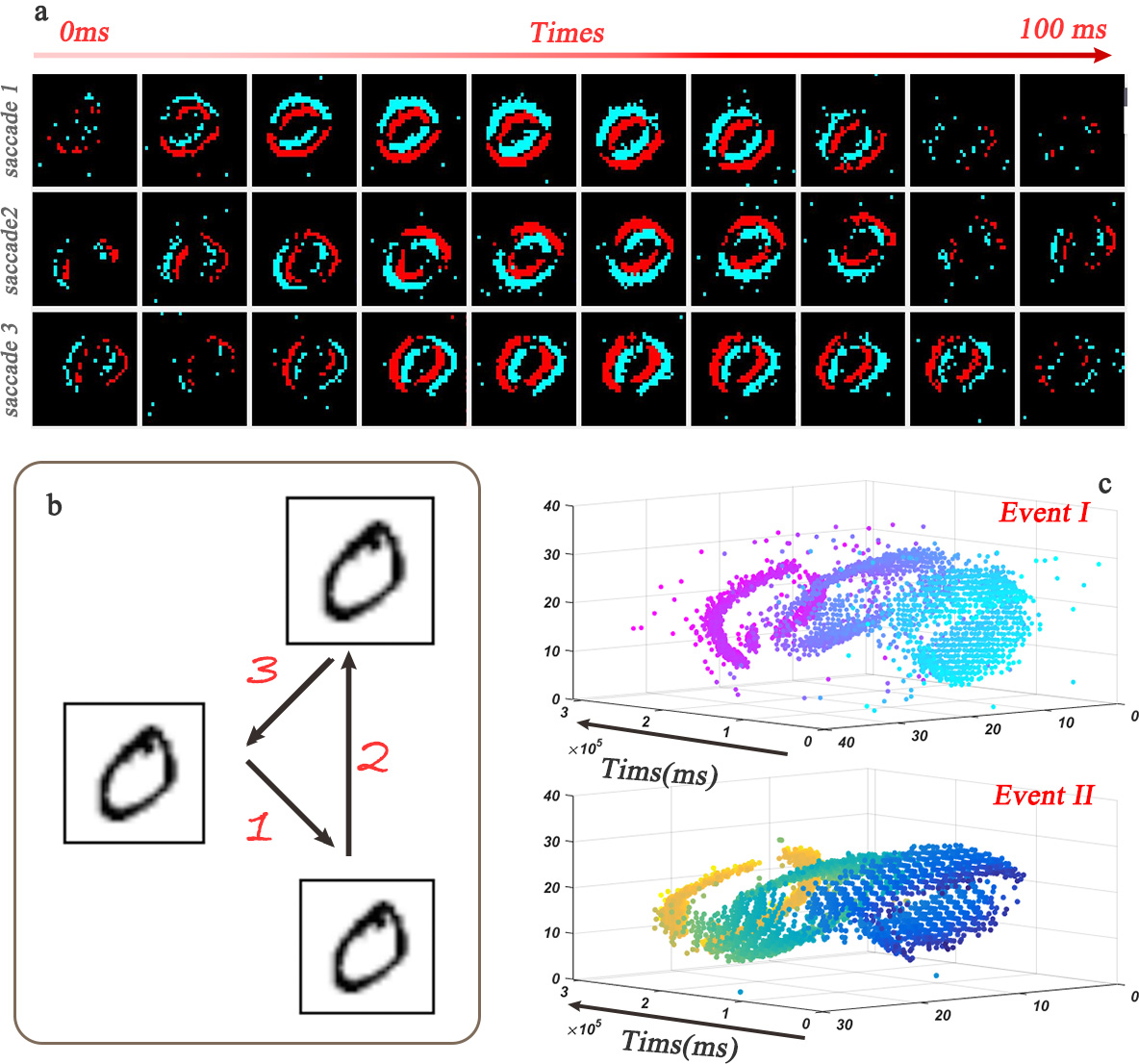}
\caption{\textbf{Dynamic dataset of N-MNIST.} (a) Each sub-picture shows a 10ms-width spike train during the saccades. (b) Spike train is generated by moving the dynamic vision sensor (DVS) in turn towards the direction of 1, 2 and 3. (c) Spatio-temporal representation of the spike train from digit 0  \cite{Orchard2015Converting}where the upper one and lower one denote the on-events and off-events, respectively.}
\label{fig5}
\end{figure}

\textbf{Dynamic Dataset.} Compared with the static dataset, dynamic dataset, such as the N-MNIST\cite{Orchard2015Converting}, contains richer temporal features, and therefore it is more suitable to exploit SNN's potential ability. We use the N-MNIST database as an example to evaluate the capability of our STBP method on dynamic dataset. N-MNIST converts the mentioned static MNIST dataset into its dynamic version of spike train by using the dynamic vision sensor (DVS)\cite{Lichtsteiner2007A}. For each original sample from MNIST, the work \cite{Orchard2015Converting} controls the DVS to move in the direction of three sides of the isosceles triangle in turn (figure\ref{fig5}.b) and collects the generated spike train which is triggered by the intensity change at each pixel. Figure\ref{fig5}.a records the saccade results on digit 0. Each sub-graph records the spike train within 10ms and each 100ms represents one saccade period. Due to the two possible change directions of each pixel intensity (brighter or darker), DVS could capture the corresponding two kinds of spike events, denoted by on-event and off-event, respectively (figure\ref{fig5}.c). Since N-MNIST allows the relative shift of images during the saccade process, it produces 34$\times$34 pixel range. And from the spatio-temporal representation in figure\ref{fig5}.c, we can see that the on-events and off-events are so different that we use two channel to distinguish it. Therefore, the network structure is 34$\times$34$\times$2-400-400-10.\\

\begin{table*}[!ht]
\centering
\caption{Comparison with state-of-the-art networks over N-MNIST.}
\label{table4}
\scalebox{1}{
\begin{tabular}{cccc}
\hline
Model & Network structure        & Training skills    & Accuracy \\ \hline
Non-spiking CNN(BP)\cite{Neil2016Phased}     & -              & None             & 95.30\%          \\
Non-spiking CNN(BP)\cite{Neil2016Effective}     & -              & None        & 98.30\%          \\
Non-spiking MLP(BP)\cite{Lee2016Training}     & $34\times34\times2$-800-10   & None             & 97.80\%          \\
LSTM(BPTT)\cite{Neil2016Phased}     & -             & Batch normalization & 97.05\%          \\
Phased-LSTM(BPTT)\cite{Neil2016Phased}     & -      & None             & 97.38\%          \\ \hline
Spiking CNN(pre-training*)\cite{Neil2016Effective}     & -              & None        & 95.72\%          \\
Spiking MLP(BP)\cite{Lee2016Training}     & $34\times34\times2$-800-10     & \begin{tabular}{c}  Error normalization/\\parameter regularization  \end{tabular}
    & 98.74\%          \\
Spiking MLP(BP)\cite{Cohen2016Skimming}     & $34\times34\times2$-10000-10  & None            & 92.87\%          \\
Spiking MLP(STBP)     & $34\times34\times2$-800-10          & None             & \textbf{98.78\%} \\ \hline
\end{tabular}}
\begin{tablenotes}
\item[1]\footnotesize{We only show the network structure based on MLP, and the other network structure refers to the above references. *means that their model is based on pre-trained ANN models.}
\end{tablenotes}
\end{table*}
Table\ref{table4} compares our STBP method with some state-of-the-art results on N-MNIST dataset.  The upper 5 results are based on ANNs, and lower 4 results including our method uses SNNs.
The ANNs methods usually adopt a frame-based method, which collects the spike events in a time interval ($50ms\sim300ms$) to form a frame of image, and use the conventional algorithms for image classification to train the networks. Since the transformed images are often blurred, the frame-based preprocessing is harmful for model performance and  abandons the hardware friendly event-driven paradigm. As can be seen from Table\ref{table4}, the models of ANN are generally worsen than the models of SNNs.
In contrast, SNNs could naturally handle event stream patterns, and by better use of spatio-temporal feature of event streams, our proposed STBP method achieves best accuracy of 98.78\% when compared all the reported ANNs and SNNs methods. The greatest advantage of our method is that we did not use any complex training skills, which is beneficial for future hardware implementation.

\subsubsection{Spatio-temporal convolution neural network}
Extending our framework to convolution neural network structure allows the network going deeper and grants network more powerful SD information. Here we use our framework to establish the spatio-temporal convolution neural network.
Compared with our spatio-temporal fully connected network, the main difference is the processing of the input image, where we use the convolution in place of the weighted summation. Specifically, in the convolution layer, each convolution neuron receives the convoluted input and updates its state according to the LIF model. In the pooling layer, because the binary coding of SNNs is inappropriate for standard max pooling, we use the average pooling instead.

\begin{table*}[!htp]
\centering
\caption{Comparison with other spiking CNN over MNIST.}
\begin{tabular}{lll}
\hline
Model                         & Network structure            & Accuracy \\ \hline
Spiking CNN (pre-training$^*$)\cite{Esser2016Convolutional}               & 28$\times$28$\times$1-12C5-P2-64C5-P2-10     & 99.12\%  \\
Spiking CNN(BP)\cite{Lee2016Training}                 & 28$\times$28$\times$1-20C5-P2-50C5-P2-200-10 & 99.31\%  \\
Spiking CNN (STBP)     & 28$\times$28$\times$1-15C5-P2-40C5-P2-300-10 & \textbf{99.42\%}  \\ \hline
\end{tabular}
\begin{tablenotes}
\item[1] \footnotesize{We mainly compare with these methods that have the similar network architecture, and * means that their model is based on pre-trained ANN models.}
\end{tablenotes}
\label{table8}
\end{table*}

\begin{table*}[!htb]
\centering
\caption{Comparison with the typical CNN over object detection dataset.}
\scalebox{1}{
\begin{tabular}{cccc}
\hline
Model     & Network structure & \multicolumn{2}{c}{Accuracy} \\ \cline{3-4}
          &                   & Mean   & Interval$^{*}$       \\ \hline
Non-spiking CNN(BP)        &  $28\times28\times1$-6C3-300-10                 & 98.57\% & [98.57\%, 98.57\%]  \\
Spiking CNN(STBP)  & $28\times28\times1$-6C3-300-10     & \textbf{98.59\%} & [\textbf{98.26\%}, \textbf{98.89\%}] \\\hline
\end{tabular}}
\begin{tablenotes}
\item[1] \footnotesize{* results with epochs [201,210].}
\end{tablenotes}
\label{table7}
\end{table*}

Our spiking CNN model are also tested on the MNIST dataset as well as the object detection dataset . In the MNIST, our network contains one convolution layers with  kernel size of $5\times5$ and two average pooling layers alternatively, followed by one hidden layer. And like traditional CNN, we use the elastic distortion \cite{Simard2003Best} to preprocess dataset. Table\ref{table8} records the state-of-the-art performance spiking convolution neural networks over MNIST dataset. Our proposed spiking CNN model obtain 98.42\% accuracy, which outperforms other reported spiking networks with slightly lighter structure. Furthermore, we configure the same network structure on a custom object detection database to evaluate the proposed model performance.  The testing accuracy is reported after training 200 epochs. Table\ref{table7} indicates our spiking CNN model could achieve a competitive performance with the non-spiking CNN.

\subsection{Performance Analysis}
\label{section-3.3}
\subsubsection{The Impact of Derivative Approximation Curves}

\begin{figure*}[!ht]
\centering
\includegraphics[width=10cm]{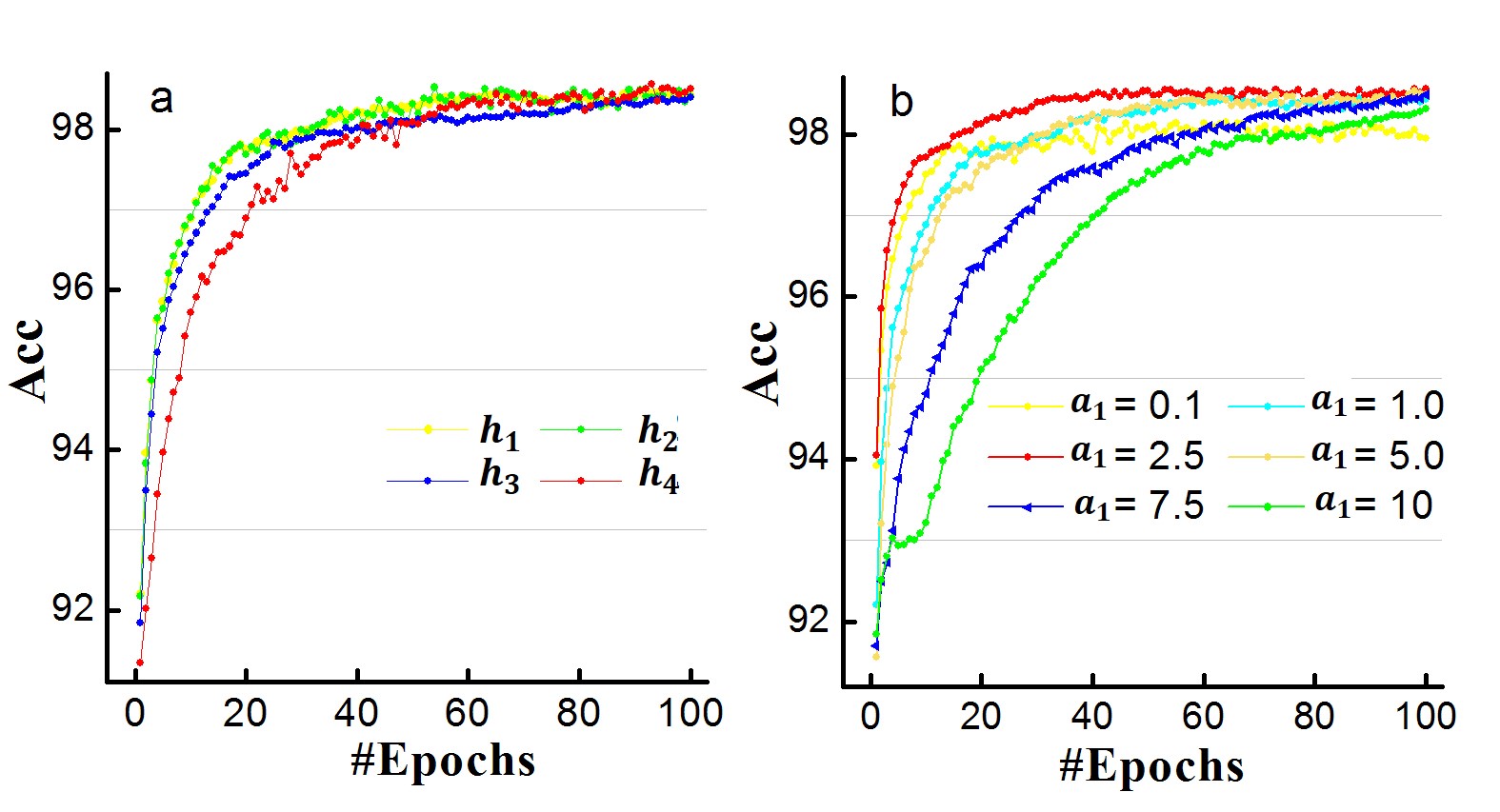}
\caption{\textbf{Comparisons of different derivation approximation curves.} (a) The impact of different approximations. (b) The impact of different widths of regular approximation.}
\label{fig6}
\end{figure*}

In section \ref{section:2.3}, we introduce different curves to approximate the ideal derivative of the spike activity. Here we try to analyze the influence of different approximation curves on the testing accuracy. The experiments are also conducted on the MNIST dataset, and the network structure is $784-400-10$. The testing accuracy is reported after training 200 epochs. Firstly, we compare the impact of different curve shapes on model performance. In our simulation we use the mentioned $h_1$, $h_2$, $h_3$ and $h_4$ shown in Figure\ref{fig3}.b. Figure\ref{fig6}.a illustrates the results of approximations of different shapes. We observe that different nonlinear curves, such as $h_1$, $h_2$, $h_3$ and $h_4$, only present small variations on the performance.

Furthermore, we use the rectangular approximation as an example to explore the impact of width on the experiment results. We set $a_1=0.1,1.0, 2.5, 5.0,7.5, 10$ and corresponding results are plotted in figure\ref{fig6}.b.  Different colors denote different $a_1$ values. Both too large and too small $a_1$ value would cause worse performance and in our simulation, $a_1=2.5$ achieves the highest testing accuracy, which implies the width and steepness of rectangle influence the model performance. Combining figure \ref{fig6}.a and figure \ref{fig6}.b, it indicates that the key point for approximating the derivation of the spike activity is to capture the nonlinear nature, while the specific shape is not so critical.\\

\subsubsection{The Impact of Temporal Domain}
A major contribution of this work is introducing the temporal domain into the existing spatial domain based BP training method, which makes full use of the spatio-temporal dynamics of SNNs and enables the high-performance training. Now we quantitatively analyze the impact of the TD item.
The experiment configurations keep the same with the previous section ($784-400-10$) and we also report the testing results after training 200 epochs. Here the existing BP in the SD is termed as SDBP.

\begin{table*}[!ht]
\centering
\caption{Comparison for the SDBP model and the STBP model on different datasets.}
\label{table6}
{
\begin{tabular}{cccccc}
\hline
Model     & Dataset            & Network structure & Training skills & \multicolumn{2}{c}{Accuracy}             \\ \cline{5-6}
          &                    &        &  &Mean    & Interval$^*$                   \\ \hline
Spiking MLP & Objective tracking &  784-400-10      & None   & 97.11\% & [96.04\%,97.78\%]  \\
(SDBP) & MNIST              & 784-400-10   &  None     & 98.29\% & [98.23\%, 98.39\%] \\ \hline
Spiking MLP & Objective tracking &  784-400-10      & None   & \textbf{98.32\%} & [\textbf{97.94\%}, \textbf{98.57\%}]  \\
(STBP)  & MNIST              & 784-400-10   & None       & \textbf{98.48\%} & [\textbf{98.42\%}, \textbf{98.51\%}] \\ \hline
\end{tabular}}
\begin{tablenotes}
\item[1] \footnotesize{* results with epochs [201,210].}
\end{tablenotes}
\end{table*}

Table\ref{table6} records the simulation results. The testing accuracy of SDBP is lower than the accuracy of the STBP on different dataset, which shows the time information is beneficial for model performance. Specifically, compared to the STBP, the SDBP has a 1.21\% loss of accuracy on the objective tracking dataset, which is 5 times larger than the loss on the MNIST. And results also imply that the performance of SDBP is not stable enough. In addition to the interference of the dataset itself, the reason for this variation may be the unstability of SNNs training. Actually, the training of SNNs relies heavily on the parameter initialization, which is also a great challenge for SNNs applications. In many reported works, researchers usually leverage some special skills or mechanisms to improve the training performance, such as the lateral inhibition, regularization, normalization, etc.
In contrast, by using our STBP training method, much higher performance can be achieved on the same network.
Specifically, the testing accuracy of STBP reaches 98.48\% on MNIST and 98.32\% on the object detection dataset. Note that the STBP can achieve high accuracy without using any complex training skills. This stability and robustness indicate that the dynamics in the TD fundamentally includes great potential for the SNNs computing and this work indeed provides a new idea.

\section{Conclusion}
In this work, a unified framework that allows supervised training spiking neural networks   just like implementing backpropagation in deep neural networks (DNNs) has been built by exploiting the spatio-temporal information in the networks.
Our major contributions are summarized as follows:
 \begin{enumerate}
 \item We have presented a framework based on an iterative leaky integrate-and-fire model, which enables us to implement spatio-temporal backpropagation on SNNs. Unlike previous methods primarily focused on its spatial domain features, our framework further combines and exploits the features of SNNs in both the spatial domain and temporal domain;

  \item  We have designed the STBP training algorithm and implemented it on both MLP and CNN architectures. The STBP has been verified on both static and dynamic datasets.
      Results have shown that our model is superior to the state-of-the-art SNNs on relatively small-scale networks of spiking MLP and CNNs, and outperforms DNNs with the same network size on dynamic N-MNIST dataset.
      An attractive advantage of our algorithm is that it doesn't need extra training techniques which generally required by existing schemes, and is easier to be implemented in large-scale networks. Results also have revealed that the use of spatio-temporal complexity to solve problems could fulfill the potential of SNNs better;

   \item  We have introduced an approximated derivative to address the non-differentiable issue of the spike activity. Controlled experiment indicates that the steepness and width of approximation curve would affect the model's performance and the key point for approximations is to capture the nonlinear nature, while the specific shape is not so critical.\\

\end{enumerate}
Because the brain combines complexity in the temporal and spatial domains to handle input information, we also would like to claim that implementing STBP on SNNs is more bio-plausible than applying BP on DNNs.
The property of STBP that doesn't rely on too many training skills makes it more hardware-friendly and useful for the design of neuromorphic chip with online learning ability.
Regarding the future research topics, two issues we believe are quite necessary and very important. One is to apply our framework to tackle more problems with the timing characteristics, such as dynamic data processing, video stream identification and speech recognition. The other is how to accelerate the supervised training of large scale SNNs based on GPUs/CPUs or neuromorphic chips. The former aims to further exploit the rich spatio-temporal features of SNNs to deal with dynamic problems, and the later may greatly prompt the applications of large scale of SNNs in real life scenarios.

\ifCLASSOPTIONcaptionsoff
  \newpage
\fi



\bibliographystyle{IEEEtran}
\bibliography{SNNLSTM}
%
%
%

%
\end{document}